\documentclass{article}

\usepackage{microtype}
\usepackage{graphicx}
\usepackage{subfigure}
\usepackage{booktabs} % for professional tables

\usepackage{hyperref}

\usepackage[accepted]{icml2024}

\usepackage{amsmath}
\usepackage{amssymb}
\usepackage{mathtools}
\usepackage{amsthm}
\usepackage{enumitem}
\usepackage{paralist}
\usepackage{xcolor}
\usepackage[utf8]{inputenc}
\usepackage{xspace}

\newcommand{\arxiv}{\textsc{Ogbn-arxiv}\xspace}
\newcommand{\arxivyear}{\textsc{Arxiv-year}\xspace}

\usepackage{multirow}

\usepackage[capitalize,noabbrev]{cleveref}

\theoremstyle{plain}

\theoremstyle{definition}

\theoremstyle{remark}

\usepackage[textsize=tiny]{todonotes}

\begin{document}

\twocolumn[
\icmltitle{Position: Graph Foundation Models are Already Here}



\icmlsetsymbol{equal}{*}

\begin{icmlauthorlist}
\icmlauthor{Haitao Mao}{equal,first}
\icmlauthor{Zhikai Chen}{equal,first}
\icmlauthor{Wenzhuo Tang}{first}
\icmlauthor{Jianan Zhao}{second,third}
\icmlauthor{Yao Ma}{fourth}
\icmlauthor{Tong Zhao}{fifth}
\icmlauthor{Neil Shah}{fifth}
\icmlauthor{Mikhail Galkin}{sixth}
\icmlauthor{Jiliang Tang}{first}
\end{icmlauthorlist}

\icmlaffiliation{first}{Michigan State University, East Lansing, US}
\icmlaffiliation{second}{Universit\'e de Montr\'eal}
\icmlaffiliation{third}{Mila - Qu\'{e}bec AI Institute}
\icmlaffiliation{fourth}{Rensselaer Polytechnic Institute, Albany, US}
\icmlaffiliation{fifth}{Snap Inc., USA.}
\icmlaffiliation{sixth}{Intel Labs}

\icmlcorrespondingauthor{Haitao Mao}{haitaoma@msu.edu}
\icmlcorrespondingauthor{Zhikai Chen}{chenzh85@msu.edu}

\icmlkeywords{Machine Learning, ICML}

\vskip 0.3in
]



\printAffiliationsAndNotice{\icmlEqualContribution} % otherwise use the standard text.

\begin{abstract}
Graph Foundation Models (GFMs) are emerging as a significant research topic in the graph domain, aiming to develop graph models trained on extensive and diverse data to enhance their applicability across various tasks and domains. 
% \new{that benefits from training on broad data, giving the potential to apply on different tasks and graphs in different domains.} 
Developing GFMs presents unique challenges over traditional Graph Neural Networks (GNNs),  which are typically trained from scratch for specific tasks on particular datasets
The primary challenge in constructing GFMs lies in effectively leveraging vast and diverse graph data to achieve positive transfer. 
Drawing inspiration from existing foundation models in the CV and NLP domains, we propose a novel perspective for the GFM development by advocating for a ``graph vocabulary'', in which the basic transferable units underlying graphs encode the invariance on graphs. We ground the graph vocabulary construction from essential aspects including network analysis, expressiveness, and stability. Such a vocabulary perspective can potentially advance the future GFM design in line with the neural scaling laws. All relevant resources for GFMs design can be found at \href{https://github.com/CurryTang/Towards-Graph-Foundation-Models-New-perspective-}{here}.
\end{abstract}

\section{Introduction}

Foundation models~\cite{bommasani2021opportunities}, which are pre-trained on massive data and can be adapted to tackle a wide range of downstream tasks, have achieved inimitable success in various domains, e.g., computer vision (CV)~\cite{CLIP} and natural language processing~(NLP)~\cite{gpt4, touvron2023llama}. 
Typically, foundation models can effectively utilize both the prior knowledge obtained from the pre-training stage and the data from downstream tasks to achieve better performance~\cite{han2021pre} and even deliver promising efficacy with few-shot task demonstrations~\cite{icl, mao2024data}. 

Meanwhile, graphs are vital and distinctive data structures that encapsulate non-Euclidean and intricate object relationships. 
Since various graphs embody unique relations, most graph learning approaches are tailored to train from scratch for a single task on a particular graph. 
This approach necessitates separate data collection and deployment for each individual graph and task. 
Consequently, an intriguing question emerges: \textit{Is it possible to devise a Graph Foundation Model~(GFM) that benefiting from large-scale training with better generalization across different domains and tasks?}

Despite advanced foundation models in other domains, the development of GFMs remains in the infant stage. 
Recent research has demonstrated initial successes of GFMs in specialized areas, such as knowledge graphs~\cite{galkin2023towardsULTRA, galkin2024zero} and molecules~\cite{beaini2023towardsMoleculeFM}. 
Notably, most of these models are built on principles specific to their domains. 
For instance, ULTRA for knowledge graph completion~\cite{galkin2023towardsULTRA} draws inspiration from double equivariance for inductive link prediction~\cite{gao2023doubleEquivariant}.
However, there is still a lack of general guidance on how to build GFMs that can effectively cater to a broad spectrum of graph-based applications.

The key difficulty in designing GFMs lies in finding the invariance across diverse graph data, ranging from social networks to molecular graphs with countless structural patterns, into the same representation space to achieve positive transfer.
% The key difficulty in designing GFMs is how to \new{align} 
The answer from the CV and NLP domains is a shared vocabulary. 
In the NLP foundation models, the text is first broken down into smaller units based on the vocabulary, which can be words, phrases, or symbols. 
In the CV foundation models, the image is mapped to a series of discrete image tokens~\cite{yu2023language, bai2023sequential} based on the vision token vocabulary. 
The vocabulary defines the basic units in the particular domain, transferable across different tasks and datasets.  
Therefore, the key challenges in achieving the GFM narrow down to how we can find the \textbf{graph vocabulary}, the basic transferable units underlying graphs to encode the invariance on graphs. 

However, finding a suitable graph vocabulary that works across diverse graphs is challenging, which is the primary focus of this paper.

\textbf{Our contributions: }
In this paper, we present a vocabulary perspective to clearly state the position of the GFM. 
In particular, we attribute the existing success of primitive GFMs to the suitable vocabulary construction guided by the particular transferability principle on graphs in Section~\ref{sec:definition}. 
A comprehensive review of the graph transferability principles and corresponding actionable steps is illustrated in Section~\ref{sec:principle}, serving us the principle for future vocabulary construction and the GFM design. 
In Section~\ref{sec:neural-scaling}, we discuss the potential for building the GFM following neural scaling laws from several perspectives (1) building and training vocabulary from scratch, and (2) leveraging existing LLM. 
Finally, we introduce more insights and open questions to inspire constructive discussions on the GFM in Section~\ref{sec:next}.

\section{Existing GFMs and Key Designs \label{sec:definition}  }

Existing GFMs~\cite{galkin2023towardsULTRA, zheng2023predicting} have achieved initial success, including promising zero-shot generalization to unseen graphs.
Based on model transferability, current GFMs can be categorized into task-specific, domain-specific, and primitive GFMs. Definitions for all categories can be found in Section~\ref{sec:pgfm}. 
The key to a successful GFM design is further discussed in Section~\ref{subsec:existing-success}. 
Notably, none of the current GFM have the capability to transfer across all graph tasks and datasets from all domains, despite such expectations being achieved in the NLP domain~\cite{gpt4, touvron2023llama} with long-term effort. GFMs remain in a nascent stage with limited development. 
Despite the gap compared to the success in the NLP domain, GFMs have already achieved significant improvement over existing GNNs with end-to-end training on a single dataset. 
However, the feasibility of general GFM remains unclear with unique graph challenges. 
Graphs are abstract data structures which are more diverse than natural language text and images grounded in the physical world

\subsection{Existing GFM Categories}
\label{sec:pgfm}
Based on the model transferability across domains and tasks, we can roughly distinguish the existing primitive GFMs into three categories: task-specific, domain-specific, and primitive GFMs. 
We provide definitions and examples for each category, with a more comprehensive illustration in Appendix~\ref{app:exist-gfm}.

A \textit{task-specific/domain-specific GFM} should be transferable across the specific task/domain and thus adapt to diverse downstream datasets and domain-specific tasks. 
A notable example of a task-specific GFM is ULTRA~\citep{galkin2023towardsULTRA}, achieving superior zero-shot knowledge graph completion performance across datasets from various domains. 
A task-specific GFM shows great practical benefits, as it can be trained on data-rich domains, e.g., Wikipedia knowledge graphs, and subsequently improve effectiveness in resource-limited domains, e.g., geography knowledge graph. 
A domain-specific GFM instance, DiG~\cite{zheng2023predicting}, learns universal representations across various chemical tasks by leveraging domain-specific knowledge. 
The domain-specific GFM is highly efficient, as one model can serve all tasks while also delivering improved effectiveness compared to single-task models.

A \textit{primitive GFM} exhibits the capability to generalize towards a limited number of datasets and tasks. 
A notable example is OFA~\cite{oneforall}, which is co-trained on data ranging from citation networks and molecule graphs to knowledge graphs via a unified task formation on node, link, and graph level tasks. The OFA model can achieve comparable or even better performance over the vanilla GNNs on each task. Nonetheless, OFA requires transforming all node features into text for co-training, which may not be convenient for all types of data. This co-training paradigm may also limit its generalization to unseen tasks and domains.

\subsection{The Key to A Successful GFM Design.\label{subsec:existing-success}}
Despite the empirical success achieved by existing GFMs, most of them are inspired by domain/task-specific principles. In this section, we aim to illustrate the common design approach using ULTRA~\cite{galkin2023towardsULTRA} as a showcase. 

ULTRA~\cite{galkin2023towardsULTRA} is a task-specific GFM focusing on the knowledge graph completion~(KGC) task. 
The KGC task aims to infer the missing triplet~(edge), denoted as $(h, r, t)$, where $r$ is a query relationship, $h$ and $t$ are the head and tail entities, respectively. 
The KGC model aims to answer the query $(h, r, ?)$ by predicting the tail entity $t$. 

The first reason for its success is to utilize the NBFNet~\cite{zhu2021neuralNBF} backbone model which enables the inductive generalization to new graphs with an expressive relational vocabulary. 
The NBFNet proposes a conditional message passing that can learn the pairwise-node representation conditioned on a head entity node and a query relation.

\citet{huang2023theoryLP} demonstrates that this conditional message passing, grounded in the relational Weisfeiler-Leman algorithm, theoretically offers greater expressiveness in KGC  compared to standard, unconditional GNNs~\cite{li2022graph}.
Such expressiveness helps to distinguish the difference between knowledge graphs with different structural features, leading to a suitable relational vocabulary.  
In contrast, \citet{barcelo2022weisfeiler} indicates that those unconditional GNNs, e.g., R-GCN~\cite{schlichtkrull2018modeling} and CompGCN~\cite{vashishth2019composition}, map non-isomorphic node pairs into the same representation, leading to a contracted relational vocabulary. 
Such contracted vocabulary may lead to negative transfer with inappropriately generalizing knowledge across non-isomorphic node pairs with inherent differences.

However, such expressive relational vocabulary only considers the pre-defined relation types which cannot generalize to the scenario with new relation types during inference. 
To extend the existing relational vocabulary including new relationship type, \citet{galkin2023towardsULTRA} constructs a graph of relations that captures fundamental interactions independent from any graph-specific relation types, serving as the second reason for its success.
The graph of relations is theoretically grounded~\cite {gao2023double} which aims to learn the double permutation-equivariant representations. 
Such representation is equivariant to permutations of both node entities and edge relation types. 
Such equivariance can be an analogy to a shared relational vocabulary. 
It connects the new unseen relationship types to the existing ones and maps the equivariant node pairs into the same representation despite different relation types, leading to the positive transfer.

In summary, we can conclude the key for ULTRA to achieve good transferability is finding a suitable vocabulary for KGC satisfying two principles: 
(1) The vocabulary should not be compacted, which causes distinct node pairs to share representations, leading to potential negative transfer.
(2) The vocabulary should be sufficiently inclusive to map new, unknown relationships onto the existing vocabulary, potentially enabling positive transfer. 
Notably, the vocabulary design in GFMs does not necessarily correspond to a tokenizer or an embedding layer as in the NLP domain. 
Instead, it can involve a model that maps graphs from different domains into the same representation space, enabling positive transfer and serving as a prerequisite for data-scaling.

The effectiveness of finding a suitable vocabulary for building the GFM can also be found in other existing primitive GFMs with the following evidence.
GraphGPT~\cite{zhao2023graphgpt} constructs a dataset-specific vocabulary where each node corresponds to a unique node ID. Notably, GraphGPT requires specific pre-training and fine-tuning on each dataset. 
MoleBERT~\cite{xia2023molebert}, the foundation model for molecule graphs, manually designs a vocabulary that transforms atom attributes into chemically meaningful codes.

\section{Graph Transferability Principles with Actionable Steps \label{sec:principle}}  
% underlying

In the last section, we investigate the key to building an effective GFM, which lies in constructing a suitable graph vocabulary to keep the essential invariance across datasets and tasks. 
Despite existing successes, more graph transferability principles, identifying different invariances, can serve as guidance for constructing new suitable graph vocabulary for future GFMs. 
We present a few actionable next steps inspired by these principles, highlighting their potential benefits.

The following discussions are organized as follows: We first provide a general introduction to the graph transferability principles in Section~\ref{sec:gprinciple}. 
Detailed task-specific principles on node classification, link prediction, and graph classification tasks can be found in Section~\ref{sec:nc},~\ref{sec:lp}, and~\ref{sec:gc}, respectively. 
We finally discuss the principles for task transferability in Section~\ref{sec:taskt}.
Notably, the following discussions majorly concentrate on the transferability of the graph structure. 
The discussion about techniques for aligning the feature space can be found in Appendix~\ref{app:engineer}.

\subsection{An overview on Graph transferability principles}
\label{sec:gprinciple}

In this subsection, we introduce principles that enable transferability on graphs, focusing on three key aspects: network analysis, expressiveness, and stability. More discussion on other principles revolving on deeper GNNs can be found in Appendix~\ref{app:principle}.

\textbf{Network analysis} provides a conventional understanding of the network system by identifying fundamental graph patterns, e.g., network motif~\cite{Menczer_Fortunato_Davis_2020} and establishing the key principles, e.g., triadic closure principle~\cite{huang2015triadic} and homophily principle, which are generally valid across different domains. 
Those principles have been generally utilized to guide the design of advanced GNNs. For example, the state-of-the-art GNN for link prediction~\cite{wang2023neural} is a Neural Common Neighbor, inspired by the triadic closure principle.
Despite its effectiveness, network analysis heavily relies on expert knowledge without a provable guarantee.

\textbf{Expressiveness} provides a theoretical background as to which functions graph neural architectures can model in general, e.g., a well-known connection that graph-level performance of GNNs is bounded by Weisfeiler-Leman tests~\citep{xu2018how,morris2019weisfeiler,JMLR:v24:22-0240_WLStory}. 
The most-expressive structural representation~\cite{srinivasan2019equivalence} is the key concept describing that the representation of two node sets should be invariant if and only if the node sets are symmetric with a permutation equivalence.
Such most-expressive structural representation serves as an important principle to design a suitable graph vocabulary that perfectly distinguishes all non-isomorphic structural patterns in multi-ary prediction tasks.

\textbf{Stability}~\cite{gnn_stability} assesses the representation sensitivity to graph perturbations.  
It aims to maintain a bounded gap in predictions for pairs under minor perturbations, rather than the expressiveness only distinguishing between isomorphic and non-isomorphic cases. 
The stability imposes a stricter constraint leading to better generalization.
It can be an analogy to the constraint on the graph vocabulary where similar structure patterns should have similar representation.

\subsection{Transferability Principles in Node Classification}
\label{sec:nc}

\textbf{Network analysis.} \emph{Homophily}~\cite{khanam2020homophily}, which describes the phenomenon of linked nodes often sharing similar features~(``birds of a feather flock together''), is a longstanding principle in social science. 
It serves as the principle guidance for methods ranging from conventional pagerank~\cite{chien2021adaptiveGPRGNN} and label propagation~\cite{chawla2005learningLP} to the recent advanced GNNs.  
Existing GNN architectures, often crafted based on the homophily principle, demonstrate strong performance on diverse homophilous graphs across various domains. This adherence to homophily not only enhances model effectiveness but also facilitates model transferability among homophilous graph datasets. Notably, successful transfers among such graphs are evidenced in~\citet{ying2018graphPINSAGE}.

While homophily predominates in network analysis, it is not a universal rule. 
In many real-world scenarios, ``opposites attract", resulting in networks characterized by heterophily—where nodes are more likely to link with dissimilar nodes. 
GNNs built with the homophily principle often struggle with heterophilious networks, except in cases of ``good heterophily''~\cite{ma2021homophily,luan2021heterophily}, where GNNs can identify and leverage consistent patterns in connections between dissimilar nodes. 
However, most heterophilious networks are complex and varied, posing challenges for GNNs due to their irregular and intricate interaction patterns~\cite{luan2023graph, wang2024understanding, mao2023demystifyingNC}. 
Consequently, GNNs' transferability, more assured in homophilous graphs, is facing significant challenges in heterophilous ones.

\textbf{Stability.} \citet{you2023graphSpecReg} theoretically establishes the relationship between transferability and network stability, demonstrating that graph filters with enhanced spectral smoothness and a smaller maximum frequency response exhibit improved transferability in terms of node features and structure, respectively. 
In particular, spectral smoothness, characterized by the Lipschitz constant of the graph filter function of the 
corresponding GNN, indicates stability against edge perturbations.
The maximum frequency response, reflecting the highest spectral frequency after applying a graph filter~(essentially the largest eigenvalue of the Laplacian matrix), describes stability against feature perturbations. 

\textbf{Actionable steps inspired by principles.}  
\cite{mao2023demystifyingNC} illustrates the network analysis principle that a single GNN can perform well on either homophily patterns or heterophily patterns, but not both. 
This principle provides the actionable insight for GFM design, suggesting that the graph vocabulary for homophily patterns and heterophily patterns should be modeled separately. 
Consequently, the model backbone for GFMs in node classification should not rely on a single GNN, which only excels on either homophilic graphs or heterophilic graphs. 
A better architecture design choice could be (1) an adaptive GNN with different aggregation filters for homophilic and heterophilic graphs, or (2) a graph transformer without a fixed aggregation process.  

\citet{you2023graphSpecReg} designs a spectral regularization term inspired by the network stability to address the out-of-distribution problem. Adapting spectral regularization for GFMs could be a potential next step.

\subsection{Transferability Principles in Link Prediction}
\label{sec:lp}

\textbf{Network Analysis.}
Important network analysis principles \cite{mao2023revisitingLP} fall into three primary concepts including: 
(1) local structural proximity corresponding to the triadic closure principle~\cite{huang2015triadic}, where friends of friends become friends themselves. It inspires well-known conventional methods including CN, RA, AA~\cite{adamic2003friends}. 
(2) global structural proximity corresponding to the decay factor principle, where two nodes with more short paths between them have a higher probability of being connected. It inspires well-known conventional methods e.g., Simrank and Katz~\cite{katz1953new, jeh2002simrank}. 
(3) feature proximity corresponding to the homophily principle~\cite{murase2019structural} where shared beliefs and thoughts can be found in connected individuals.

These principles guide the evolution of link prediction algorithms, from basic heuristics to sophisticated GNNs~\cite{BUDDY, li2023evaluatingLP}. GNNs, inspired by these principles, perform well across diverse graphs in multiple domains. 
Moreover, \citet{zheng2023you} provides empirical evidence supporting the beneficial transferability of these guiding principles.

\begin{figure}
    \centering
    \includegraphics[width=0.35\textwidth]{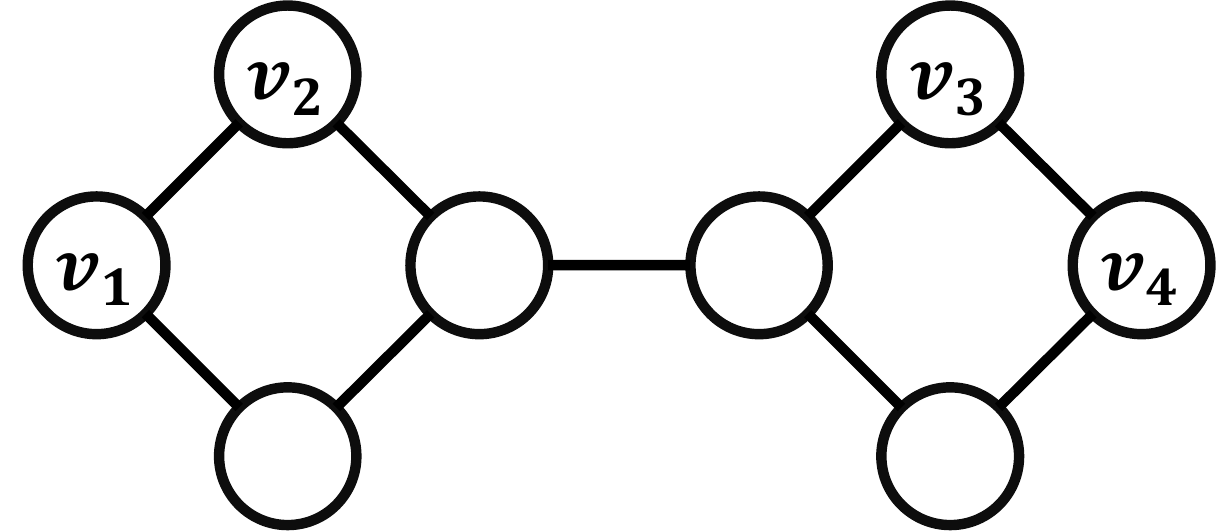}
    % \vspace{-0.8em}
    \caption{In this graph, nodes $v_1$ and $v_4$ are isomorphic; links $(v_1, v_2)$ and $(v_2, v_4)$ are not isomorphic. However, vanilla GNN with the same node representations $v_1$ and $v_4$ gives the same prediction to links $(v_1, v_2)$ and $(v_2, v_4)$.}
    \vspace{-1em}
    \label{fig:link-prediction}
\end{figure}

\textbf{Expressiveness.} 
A vanilla GNN, equipped with only single-node permutation equivalence, cannot achieve transferability for the link prediction task due to its lack of expressiveness. 
An example to showcase such failure is shown in Figure~\ref{fig:link-prediction} with a featureless graph.
$v_1$ and $v_4$ are represented identically by the vanilla GNN, as they possess identical neighborhood structures.

Therefore, the similarity between $v_1$ and $v_2$ will be the same as the one between $v_4$ and $v_2$, leading to identical representations and predictions for both links $(v_1, v_2)$ and $(v_2, v_4)$
However, according to the global structural proximity, $(v_1, v_2)$, with a shorter distance of 1, should be more likely to be connected.
The vanilla GNN, computing $v_1$'s representation solely from its neighborhood, overlooks the structural dependence with $v_2$.  
As a result, this potentially leads to negative transfer, where the GNN might erroneously predict both or neither link to exist, whereas it's more likely that only $(v_1, v_2)$ has a link.

To consider all the possible dependencies between node pairs, we aim for the most expressive structural representation for the link prediction. This representation should be invariant if and only if links are symmetric.
\citet{zhang2018link} achieves such structural representation by incorporating node labeling features that depend on both the source and target nodes in a link.
\citet{zhang2021labeling} further highlights the key aspects of node labeling design, including:
(1) target-nodes-distinguishing, where the source and target nodes have distinct labels compared to other nodes; and
(2) permutation equivariance. 
Node labeling methods that fulfill these criteria, such as double radius node labeling~(DRNL) and zero-one~(ZO) labeling, can produce the most expressive structural representations. 
Many other GNNs~\cite{you2021identity, you2019position, wang2021equivariant} can achieve similar expressiveness, serving as the potential backbone for GFM on the link prediction task. 
The expressiveness representation can find the complete set of distinct relations to differentiate all non-isomorphic node pairs, thereby mitigating the risk of negative transfer in standard GNNs.
\citet{huang2023theoryLP} extends the relational Weisfeiler-Leman framework~\citep{barcelo2022weisfeiler} to link prediction and incorporate the concept of labeling tricks to multi-relational graphs.

\textbf{Stability.} For those equally expressive structural representations, there may still be a gap in terms of their stability. For example, empirical evidence~\cite{zhang2021labeling} shows that GNNs with DRNL labeling outperform those with ZO labeling. 
From the perspective of stability, it is crucial to maintain a bounded gap in predictions for pairs under minor perturbations. 
\citet{wang2021equivariant} provides a theoretical analysis identifying key properties of stable positional encoding~(GNNs should be rotation and permutation equivariant to positional encodings) that enhance generalization. The stable positional encoding may be directly applied towards better GFMs.

\textbf{Actionable step inspired by principles.}  
\cite{mao2023revisitingLP} illustrates the network analysis principle concerning the incompatibility between structural proximities and feature proximity. 
Node pairs with high feature proximity are likely to be with low local structural proximity and vice versa. 
This incompatibility leads to over-emphasis on node pairs with high structural proximity while neglecting those with high feature proximity. 
This principle provides actionable insight for GFM design, suggesting that the graph vocabulary for feature proximity patterns and structural proximity patterns should be modeled separately. 
Consequently, the model backbone for GFMs in link prediction should separately encode the pairwise structural proximity and the feature proximity. 

A GNN following the expressiveness principles could include all the important structural information relevant to the link prediction~\cite{zhang2021labeling}. 
\citet{dong2024universal} utilizes in-context learning to effective transfer expressive GNN representations to new, unseen graphs. 
Satisfying performance can be found across graphs from biology, transport, web, and social domains. 
An actionable next step could be to better utilize expressive representations for downstream graphs from specific domains.

\subsection{Transferability Principles in Graph Classification}
\label{sec:gc}

\textbf{Network Analysis.} 
Network motifs, typically composed of small and recurrent subgraphs, are often considered the building blocks of a graph ~\cite{milo2002network, benson2016higher}. 
A proper selection of the motif set can cover most essential knowledge on the specific datasets. 
Graph kernels~\cite{vishwanathan2010graphkernel} are proposed to quantify motif counts or other pre-defined graph structural features and then utilize the extracted features to build a classifier such as SVM.
Despite the essential motif sets from different domains being generally different, there could exist a uniform set of motifs shared across different domains. 
In such cases, the positive transfer can be found on the uniform sets, where \citet{battiston2020networks} shows the positive transfer across neuronal connectivity networks, food webs, and electronic circuits. 
Therefore, we conjecture that the network motif could be the base unit for the vocabulary (a set of invariant elements) for the graph classification as it is both explainable and potentially shared across graphs.

\textbf{Expressiveness.} 
\citet{zhang2024beyond} proposes a unified framework to understand the ability of different GNNs to detect and count graph substructures~(motif).
More expressive GNN which could detect more diverse motifs and construct a richer graph vocabulary.
In analogy with the uniform motif sets, we conjecture that it is more possible for the expressive GNN to find the uniform motif sets and achieve better transferability.

\textbf{Stability.} 
\citet{huang2023stability} proposes a provably stable position encoding that surpasses the expressive sign and invariant encoding~\cite{kreuzer2021rethinking} and modeling~\cite{lim2022sign}, enabling minimal changes to positional encodings on the minor modifications to the Laplacian.
The key innovation is to apply a weighted sum of eigenvectors instead of treating each eigensubspace independently. 
Satisfying performance can be observed on the out-of-distribution molecular graph prediction. 
Such stable positional encoding may be directly applied towards better GFMs.

\textbf{Actionable step inspired by principles.}  
Inspired by the network analysis with graph kernels, one concrete next step towards GFMs could be revolving on how to identify frequent network motif~\cite{hovcevar2014combinatorial, ribeiro2021survey} which should be transferable across all graphs. Expressive GNNs with better network motif model capability could be a suitable architecture towards GFMs.

\begin{figure}[ht]
    \includegraphics[width=\linewidth]{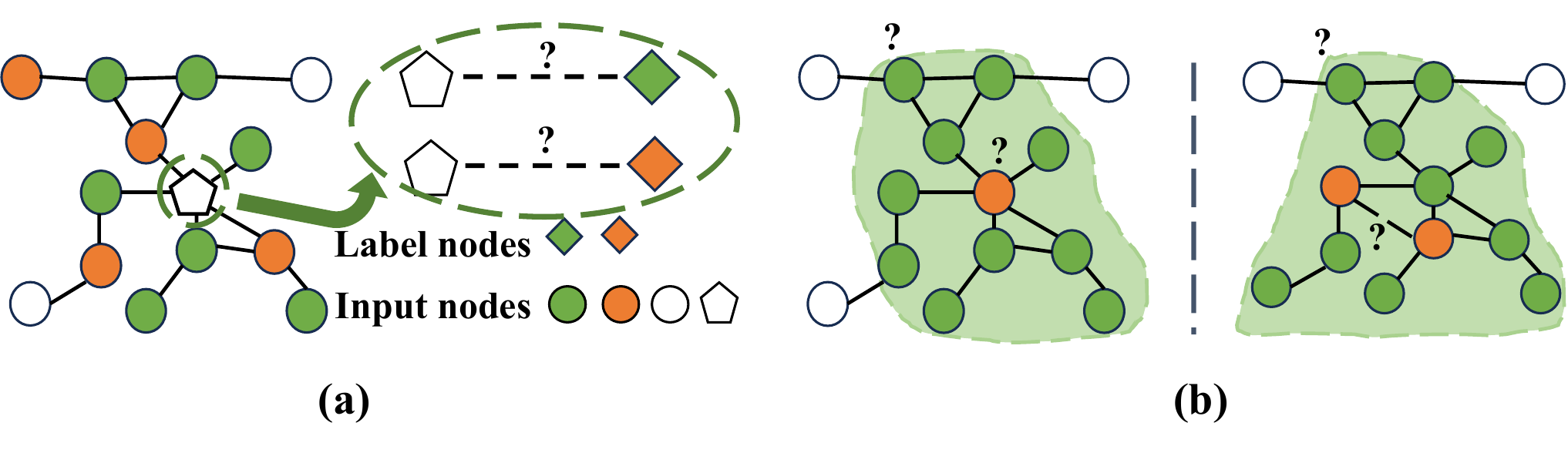}
    \vskip -0.6em
    \caption{Unifying different task formulations: (a) Link view: Given the target node, node classification is converted to the link prediction between the target node and corresponding label nodes. (b) Subgraph view: Node classification (orange node) is converted to the (green) ego-graph classification. Link prediction (orange nodes) is converted to the (green) induced-subgraph classification.}
    \label{fig:test}
    \vskip -1.3em
\end{figure}

\subsection{Transferability Principles across Tasks}
\label{sec:taskt}

A unified task formulation is generally employed to facilitate transferability across various tasks. The unified task formulation enables (1) enlarging the dataset size via converting datasets for different downstream tasks as one and (2) utilizing one pre-training model to serve different tasks. 
The significance of aligning task formulations is evident in the following example: \citet{jin2020self} shows that using link prediction directly as a pretext task leads to negative transfer for node classification. 
However, by reformulating node classification into a link prediction problem~\cite{GPPT, huang2023prodigy}, where a node's class membership is treated as the link likelihood between the node and label nodes, positive transfer is achieved. 
\citet{liu2023graphpromptWWW, sun2023all} further propose a sub-graph view to adapt the node classification as an ego graph classification, and link prediction as a binary classification on the induced sub-graph of the target node pair. 
Figure~\ref{fig:test} provides illustrative examples for these two unified views. 
More recently, \citet{oneforall} unifies node-level, link-level, and graph-level tasks via (1) adding a virtual prompt node and (2) connecting the virtual nodes to nodes of interests, i.e., the center node for node classification, source and target nodes for link prediction, and all nodes for graph classification.

A unified formulation provides the possibility for co-training all tasks together while it remains unknown whether this can be done without negative transfer. 
Moreover, the unified task formulation may not be necessary to achieve transfer across tasks. 
It is generally utilized for supervised co-training and prompt-based prediction as discussed above. 
A GFM can be (1) pre-trained with self-supervised tasks and (2) adapted to downstream tasks via fine-tuning without requiring specific task formulations. 
The success is due to the transferability principles across different tasks.
However, there remains limited study in this direction. 
We list a few existing principles as follows.
(1) Node classification and link prediction tasks share the feature homophily as an important principle. 
(2) \citet{liu2023simga} indicates that the global structural proximity principle on the link prediction can improve the node classification performance on the non-homophilous graph. 
(3) The triadic closure in the link prediction is a particular network motif utilized in the graph classification. 
There are more shared motifs~\cite{hibshman2021joint, dong2017structural, abuoda2020link, kriege2020survey} on both graph classification and link prediction tasks. 
We emphasize the importance of cross-task transferability principles as an important future direction.

\section{Neural Scaling Law on GFMs}
\label{sec:neural-scaling}

The success of the foundation model can be attributed to the validity of the neural scaling law~\cite{kaplan2020scalingNLP} which shows performance enhancement with increasing model scale and data scale. 
In this section, we first discuss when the neural scaling law happens in Section~\ref{subsec:scaling} from a graph vocabulary perspective. 
We then discuss techniques towards successful data scaling and model scaling in Section~\ref{sec: datascal} and~\ref{sec: modelscal}, respectively. 
We finally discuss the potential on leveraging large-scale LM on the graph domain in Section~\ref{subsec:scaling-llm}. 
More discussions on technical details can be found in Appendix~\ref{app:engineer}.

\subsection{When Neural Scaling Law Happens \label{subsec:scaling}} 
In section~\ref{sec:principle}, we discuss the underlying transferable principles guiding future vocabulary construction. 
Such principle guidance has led to the successful scaling behavior in the material science domain~\cite{shoghi2023jmp, zhang2023dpa2, batatia2023macemp} with the help of the geometric prior. 
Nonetheless, we are still cautious about whether the existing success can be extended to the graph domain. 
The key concern is whether graphs can strictly follow those principles. 
Uncertainty can be found on the human-defined graph construction criteria~\cite{brugere2018network}. 
For instance, the construction knowledge relying on expert knowledge may lead to uncertainty in edges~\cite{ye2022generative}. 
\citet{Chen2023ExploringTP, li2023graphcleaner} observe that the mislabeled samples widely exist across datasets, where the popular \textsc{CiteSeer} dataset has more than 15\% wrongly labeled data. 
Despite the above uncertainty, different graph constructions with manual design can follow opposite principles. 
For instance, \arxiv~\cite{hu2020openOGB} and \arxivyear~\cite{lim2021largeLINKX} are two node classification datasets with identical graph information. 
The only difference lies in the label where \arxiv employs paper categories, and \arxivyear uses publication years as labels, resulting in conflicting homophily and heterophily properties~\cite{mao2023demystifyingNC}.
Therefore, when uncertainties and opposite graph constructions exist, the scaling behavior may not happen as the data does not obey the graph transferable principles.

\subsection{Data Scaling}
\label{sec: datascal}
Data scaling refers to the phenononmon that the performance consistently improves with the increasing data scale. 
\citet{chen2023uncoveringMoleculeScaling, huang2023prodigy} initially validate that GNNs trained in both supervised and self-supervised manners follow data scaling law on molecular property predictions, and node classification on text-attributed graphs. 
\citet{cao2023preW2PGNN} further exhibits that the similarity between pre-training data and downstream task data serves as a prerequisite for the data scaling on graphs. 
Specifically, \citet{cao2023preW2PGNN} provides concrete guidance on how to select the pre-training data via the graphon signal analysis and the essential network property, i.e. network entropy, respectively. 
Notably, all principles mentioned in Section~\ref{sec:principle} can be applied to facilitate positive transfer with data scaling phenomena.

A limitation in current research on data scaling is graph data insufficiency, in contrast to the readily available trillion-level real-world data in CV and NLP domains. 
The key reasons are two-fold: (1) constructing graphs requires expert intervention e.g., defining relationships (2) intellectual property issues.
We endeavour to collect all the open-source graph datasets with details in Appendix~\ref{app:dataset}. 

Synthetic graph generation can be utilized to alleviate the data insufficiency issue, enableing more comprehensive training. 
Traditional graph generative models~\cite{albert2002statistical, robins2007introduction, airoldi2008mixed, leskovec2010kronecker} are capable of generating graphs satisfying some certain statistical properties, which still plays an important role on node-level and link-level tasks.
Deep generative models on graph~\cite{jin2020hierarchical, luo2021graphdf, jo2022score, vignac2023digress, liu2023data} have shown great success in generating high-quality synthetic graphs which helps graph-level tasks by providing a more comprehensive description of the graph distributions space. 
With successful evidence of pre-training on synthetic data from other domains~\cite{mishra2022task2sim, trinh2024solving}, we anticipate the potential on benefits from high-quality synthetic graphs.

\subsection{Model Scaling}  
\label{sec: modelscal}

Model scaling refers to the phenomenon that the performance consistently improves with the increasing model scale. 
Previous research in NLP indicates that apart from data, the backbone model constitutes a fundamental for scaling~\cite{kaplan2020scalingNLP}. 
\citet{liu2024neural} primarily validates the neural scaling law on various graph tasks and model architectures under the supervised setting. 

However, \citet{kim2022pureTOKENGT} demonstrates that the GAT~\cite{velivckovic2017graph} with a larger number of parameters underperforms on the graph regression tasks compared to the smaller-sized counterparts. 
As a comparison, geometric GNNs scale well to predict atomic potentials in material science~\cite{shoghi2023jmp, zhang2023dpa2, batatia2023macemp}. 
Observations indicate that geometric GNNs with a good geometric-prior vocabulary design can help achieve model scaling over the vanilla GNN.

Graph transformer is another popular choice for the model architecture, where geometric-prior graph vocabulary design is explicitly modeled through either a GNN encoder or positional encoding~\cite{muller2023attending}. 
\citet{gps_plusplus, unimol} show that graph transformers show positive scaling capabilities for molecular data under a supervised setting. More recently, \citet{zhao2023graphgpt} demonstrates vanilla transformer's effectiveness in protein and molecular property prediction. Particularly, it views the graph as a sequence of tokens forming an Eulerian path~\cite{Edmonds1973MatchingET}, which ensures the lossless serialization, and then adopts next-token prediction to pre-train transformers. 
After fine-tuning, it achieves promising results on the protein association prediction and molecular property prediction and shows that vanilla transformers also follow the model scaling law~\cite{kaplan2020scalingNLP}. 
Nonetheless, the effectiveness of transformers on other tasks remains unclear. 

\subsection{Leveraging Large-scale LMs for Graphs \label{subsec:scaling-llm}}

LLMs with successful scaling behavior have achieved tremendous success in the NLP domain. 
Surprisingly, well-trained LLMs can be applied to other domains with satisfying performance such as time series forecast~\cite{gruver2024large} and material science~\cite{gruver2024fine}. 
Larger-scale LLMs can even capture key symmetries of crystal structures, suggesting that LLMs may posses a strong simplicity bias~\cite{panwar2023context} across domains by implementing Bayesian model averaging algorithm~\cite{zhang2023and}. 

A recent line of research on GFMs focuses on leveraging strong capabilities of LLMs on graph tasks. 
Our discussions can be roughly categorized on LLM applications (i) conventional graph tasks (such as node, edge, and graph classification), and (ii) language-driven tasks like Graph Question Answer~(GQA).

\textbf{LLMs on conventional graph tasks.} 
One natural way to utilize LLMs is as textual feature encoders~\cite{Chen2023ExploringTP}. 
Despite original node features may not be text, \citet{oneforall} manually converts them into knowledge-enhanced text descriptions and then encodes features into textual embedding. 
This LLM embedding approach offers the following benefits. 
(i) High feature quality helps achieve satisfying performance with vanilla GCN~\cite{Chen2023ExploringTP}. 
(ii) LLMs encode diverse original features into an aligned feature space, enabling training and inference across graphs from different domains without the feature heterogeneity problem. 
Notably, when LLMs are utilized as feature encoders for textural understanding, the scaling law does not happen~\cite{behnamghader2024llm2vec}, meaning that a larger model does not necessarily lead to better performance.

Another approach is to utilize LLMs as predictors which first fine-tunes LLM and then generate predictions in a natural language form. 
\citet{Chen2023ExploringTP, he2023harnessingTAPE} treats node classification as text classification on the target node feature, illustrating promising results in the zero-shot setting.
However, simply flattening graph structures into prompts does not yield additional improvement, remaining a large performance gap compared to well-trained GNNs~\cite{Chen2023ExploringTP}. 
To better encode the graph structure knowledge, methods such as GNN~\cite{tang2023graphgpt}, graph transformer~\cite{chai2023graphllm}, and non-parametric aggregation~\cite{chen2024llaga} are utilized as structure encoders. 
The encoded structural embeddings are then linearly mapped into text space as prompt tokens. 
LLMs generate predictions based on a concatenation of the prompt token and the textual instruction. 
Instead of additional graph modeling, \citet{zhao2023graphtext} employs a novel tree-based prompt design that transforms the graph into sequence while retaining important structural semantics. 
This approach indicates the potential for LLMs to understand particular graph structures. 
Overall speaking, a proper LLM fine-tuning can achieving satisfying graph performance while the efficiency may be a potential issue.

\textbf{LLMs on language-driven graph tasks.} Instead of adapting LLM for conventional graph tasks, LLMs can also be applied to language-driven tasks they originally skilled in, for example, Graph Question Answer~(GQA).
\citet{fatemi2023talk, wang2023NLGRAPH} apply LLMs on various GQA tasks, e.g, cycle check, and maximum flow, by describing graph structure with natural language. 
More recently, \citet{perozzi2024let} incorperates an external GNN tokenizer to encode graph information, achieving satisfying out-of-domain generalization to unseen graph tasks. 
Interestingly, \citet{perozzi2024let} illustrates that equivariance is not necessary when equipped with LLMs.  
\cite{he2024g} proposes new real-world challenging GQA tasks and a corresponding LLM-based conversational framework. This framework integrates GNNs and retrieval-augmented generation~(RAG) to improve graph understanding and mitigate issues like hallucination, demonstrating effectiveness across multiple domains. 
Until now, most GQA challenges have focused on the abstract graphs without concrete descriptions for each node, creating to obstacle to o leveraging the extensive internal knowledge in LLMs. We call for more real-world GQA challengesm enabling better leverage LLM capabilities.

Despite the above successes, it remains concerns on the LLM's capability on understanding the essential graph structures. \citet{saparov2022language, dziri2023faith} theoretically observe that the LLM is required to tackle problems sequentially greedily~\cite{mccoy2023embers}, leading to a shortcut solution rather than a formal analysis on the graph structure. A more comprehensive discussion can be found in Appendix~\ref{subapp:llm-graph}.

\section{Insights \& Open Questions}\label{sec:next}
In this section, we explore key insights gained from recent advancements in GFMs and highlight open questions that remain to be addressed in this evolving field. More comprehensive discussions can be found in Appendix~\ref{app:discussion}

\subsection{Potential Redundancy on Pretext Task and Architecture Design}

There are mainly two approaches to achieving transferability: (1) designing GNNs with specific geometric properties for transfer, e.g., ULTRA~\cite{galkin2023towardsULTRA}, and (2) creating pretext tasks to automatically learn these properties. \cite{jin2020self} suggests an overlap between these approaches, indicating that pretext tasks targeting local structural information might be unnecessary, given that GNNs often inherently encode this information.
Investigating the strengths and limitations of these techniques, along with providing practical guidance for their selection, could be a valuable research direction. 
A hypothesis might be that model design methods are more suitable for data that strictly adheres to geometric priors, while pretext task designs are more effective in the opposite scenario.

\subsection{The Feasibility of GFMs} 

Graphs can be defined in different ways based on different criteria like similarity or influence between node pairs~\cite{brugere2018network}. 
We can then categorize graphs based on the observability of the criteria. 
The observable graph is unambiguously known, e.g., whether one paper cites another paper in a citation graph. 
Text and images can also be viewed as a specific case of observable graphs. 
In contrast, the unobservable ones are manually conducted with ambiguous descriptions of the relationship, e.g., whether one gene regulates the expression of another in a gene-regulate graph. 
These graphs may not naturally exist in the world, leading to uncertainty with a lack of invariant principle.  
It remains unknown whether GFMs can learn shared knowledge while avoiding manually introduced noisy patterns.

There are concerns about the benefit of training a GFM on graphs that are neither from the same domain nor share the same downstream task. 
On the one hand, it seems that training on them simultaneously shows no positive transfer benefit while increasing the risk of the negative transfer. 
On the other hand,  there may be potential undiscovered transferable patterns that could lead to success.
Therefore, we pose an open question whether there exists a universal structural representation space that can benefit all the graph tasks?

\subsection{Broader Usage of GFM}
In this paper, we majorly focus on building GFM for conventional graph-focused tasks.   
Notably, graph formulation provides the universal representation ability, which has a broader usage in other domains, e.g., scene graphs for Computer Vision~(CV)~\cite{zhai2023commonscenes,zhong2021learningSCENEGRAPH}, bipartite graphs for linear programming~\cite{chen2022representing}, and physical graphs for understanding physical mechanisms~\cite{shi2022learning}.
To emphasize more broader usage of GFMs, we illustrate the potential advantaged usage of GFMs over existing foundation models in reasoning, computer vision, and code intelligence domains domains. Details can be found as follows. 

\textbf{Reasoning.} 
\citet{ibarz2022generalist} proposes a task-specific GFM, focusing on neural algorithmic reasoning tasks. 
A strong reasoning capability can be found with effectiveness across sorting, searching, and dynamic programming tasks. 
We argue that this GFM following the theory of algorithmic alignment~\cite{xu2020can} may achieve better reasoning capability than the LLM merely relying on the textual inputs via retrieving concepts co-occur frequently in training data~\cite{prystawski2023think}.

\textbf{Computer Vision.} Scene graph is a data structure representing objects, their attributes, and the relationships between them within an image, facilitating CV tasks such as image understanding and visual reasoning. However, current research remains a naive scene graph modeling with vanilla GNNs with more emphasis on image modeling. We argue that the GFM on the scene graph may help to preserve global and local scene-object relationships~\cite{zhai2024commonscenes}, avoiding the potential conflict or redundancy between multiple objectives which frequently appears on the recent popular Sora model~\cite{videoworldsimulators2024}.   

\textbf{Code Intelligence.} Graphs, e.g., code property graph~\cite{liu2024source}, control flow graph, and program dependency graph, play an important role in code-relevant tasks, e.g., vulnerability detection~\cite{liu2024source}, fault localization~\cite{rafi2024towards}, and code search~\cite{ling2021deep}. 
Compared to sequence-based modeling with LLMs, graphs can provide a complementary perspective on the overlooked essential code attributes such as syntax, control flow, and data dependencies. However, the graph modeling remains naive with unknown transferability across different program languages.

Overall speaking, GFMs demonstrate unique value compared to foundation models in other domains. However, they are limited to applications involving graph structure data. 
An exciting future topic is how to adaptively combine GFM with other foundation models across different modality towards a powerful Artificial General Intelligence~(AGI).

\section{Conclusion\label{sec:conclusion}}

From the transferability principles of graphs, we review existing GFMs and ground their effectiveness from a vocabulary view to find a set of basic transferrable units across graphs and tasks. 
Our key perspectives can be summarized as follows: (1) Constructing a universal GFM is challenging, but domain/task-specific GFMs are approachable with the usual availability of a specific vocabulary. (2) One challenge is developing GFMs following the neural scaling law, which requires more data collection, suitable architecture design, and properly leveraging LLMs.  
This paper summarizes the current position of GFMs and challenges toward the next step, which may be a blueprint for GFMs to inspire relevant research. 
\section*{Acknowledgement} 

We want to thank Yanqiao Zhu at the University of California, Los Angeles, and Yuanqi Du at Cornell University for their constructive comments on this paper.

Haitao Mao, Zhikai Chen, Wenzhuo Tang, and Jiliang Tang are supported by the National Science Foundation (NSF) under grant numbers CNS 2246050, IIS1845081, IIS2212032, IIS2212144, IOS2107215, DUE 2234015, DRL2025244 and IOS2035472, the Army Research Office (ARO) under grant number W911NF-21-1-0198, the National Telecommunications and Information Administration (NTIA), the Home Depot, Amazon Faculty Award, JP Morgan Faculty Award, Microsoft Research, Meta, and SNAP. 
Yao Ma is supported by the National Science Foundation (NSF) under grant numbers NSF-2406648 and NSF-2406647. 

\section*{Impact Statements}
\label{sec: bi}

In this paper, we provide principle guidance for the development of graph foundation models, which can be a pivotal infrastructure empowering diverse applications like nature science and E-commerce. 
The graph foundation model may reduce the resource consumption associated with training numerous task-specific models. 
Moreover, it may substantially curtail the requirement for manual annotation, particularly in domains such as molecular property prediction. 
We anticipate that our contributions will advance the ongoing efforts aimed at developing next-generation graph foundation models with better versatility and fairness.

\bibliography{main}
\bibliographystyle{icml2024}

%%%%%%%%%%%%%%%%%%%%%%%%%%%%%%%%%%%%%%%%%%%%%%%%%%%%%%%%%%%%%%%%%%%%%%%%%%%%%%%
%%%%%%%%%%%%%%%%%%%%%%%%%%%%%%%%%%%%%%%%%%%%%%%%%%%%%%%%%%%%%%%%%%%%%%%%%%%%%%%
% APPENDIX
%%%%%%%%%%%%%%%%%%%%%%%%%%%%%%%%%%%%%%%%%%%%%%%%%%%%%%%%%%%%%%%%%%%%%%%%%%%%%%%
%%%%%%%%%%%%%%%%%%%%%%%%%%%%%%%%%%%%%%%%%%%%%%%%%%%%%%%%%%%%%%%%%%%%%%%%%%%%%%%
\newpage
\appendix
\onecolumn

\section{A Collection of Datasets to Support Pre-training \label{app:dataset}}

In this section, we show a collection of large-scale graph datasets from various fields to support pre-training massive-scale graph foundation models. We highly suggest using NetworkRepository~\cite{nr} for large-scale pretaining, which is the largest graph database presently available. Notably, burdensome pre-processing is required to clean those noisy and disordered data.

\begin{table}[htb]
\caption{A collection of datasets together with their URL and descriptions to support larger-scale pre-training}
\resizebox{\linewidth}{!}{
\begin{tabular}{@{}lll@{}}
\toprule
\textbf{Name }                & \textbf{URL}                                                                      & \textbf{Description}                                                                  \\ \midrule
\textsc{TU-Dataset~\cite{Morris+2020}}          & \url{https://chrsmrrs.github.io/datasets/}                                     & A collection of graph-level prediction datasets                              \\
\textsc{NetworkRepository~\cite{nr}}    & \url{https://networkrepository.com/}                                           & The largest graph datasets, with graphs coming from 30+ different domains    \\
\textsc{Open Graph Benchmark~\cite{hu2020openOGB}} & \url{https://ogb.stanford.edu/}                                                & Contains a bunch of large-scale graph benchmarks                               \\
\textsc{Pyg~\cite{fey2019fast}}                  & \url{https://pytorch-geometric.readthedocs.io} & Official datasets provided by \textsc{PYG}, containing popular datasets for benchmark \\
\textsc{SNAP~\cite{snapnets}}                 & \url{https://snap.stanford.edu/data/}                                          & Mainly focus on social network                                               \\
\textsc{Aminer~\cite{aminer}}               & \url{https://www.aminer.cn/data/}                                              & A collection of academic graphs                                              \\
\textsc{OAG~\cite{OAG}}                  & \url{https://www.aminer.cn/open-academic-graph}                               & A large-scale academic graph                                                 \\
\textsc{MalNet~\cite{freitas2021malnet}}               & \url{https://www.mal-net.org/\#home}                                           & A large-scale function calling graph for malware detection                   \\
\textsc{ScholKG~\cite{Dess2022CSKGAL}}              & \url{https://scholkg.kmi.open.ac.uk/}                                          & A large-scale scholarly knowledge graph                                      \\ 
\textsc{Graphium~\cite{beaini2023towardsMoleculeFM}}          & \url{https://github.com/datamol-io/graphium}                                    & A massive dataset for molecular property prediction \\
\textsc{Live Graph Lab~\cite{zhang2023live}}           & \url{https://livegraphlab.github.io/}                    & A large-scale temporal graph for NFT transactions \\
\textsc{Temporal Graph Benchmark~\cite{huang2023temporal}}   & \url{https://docs.tgb.complexdatalab.com/}        & A large-scale benchmark for temporal graph learning \\
\textsc{MoleculeNet~\cite{MoleculeNet}}               & \url{https://moleculenet.org/}                         & A benchmark for molecular machine learning \\
\textsc{Recsys data~\cite{recsys}}                  & \url{https://cseweb.ucsd.edu/~jmcauley/datasets.html}     & A collection of datasets for recommender systems \\
\textsc{LINKX~\cite{lim2021largeLINKX}}        & \url{https://github.com/CUAI/Non-Homophily-Large-Scale}                       & A collection of large-scale non-homophilous graphs \\ 
\textsc{CLRS~\cite{velivckovic2022clrs}} & \url{https://github.com/google-deepmind/clrs} & A collection of algorithmic reasoning datasets. \\ 
\textsc{GraphQA~\cite{he2024g}} & \url{https://github.com/XiaoxinHe/G-Retriever} & A collection of graph question answer datasets. \\

\bottomrule
\end{tabular}}
\end{table}

\section{Existing GFMs}
\label{app:exist-gfm}

In this section, we demonstrate existing representative GFMs and categorize them into \textit{primitive GFM}, \textit{domain-specific GFM}, and \textit{task-specific GFM}, as shown in Table~\ref{tab:gfm}.

\begin{table}[htb]
\centering
\caption{A collection of existing GFMs.}
\label{tab:gfm}
\resizebox{0.8\linewidth}{!}{
\begin{tabular}{@{}llll@{}}
\toprule
 &
  \textbf{Name} &
  \textbf{Domain} &
  \textbf{Task} \\ \midrule
\multirow{4}{*}[-1em]{\textbf{\begin{tabular}[c]{@{}l@{}}Primitive \\ GFM\end{tabular}}} &
  \textsc{PRODIGY~\cite{huang2023prodigy}} &
  \begin{tabular}[c]{@{}l@{}}Text-attributed graph, \\ Knowledge graph\end{tabular} &
  \begin{tabular}[c]{@{}l@{}}Node classification, \\ Knowledge graph reasoning\end{tabular} \\
 &
  \textsc{OneForAll~\cite{oneforall}} &
  \begin{tabular}[c]{@{}l@{}}Text-attributed graph, \\ Knowledge graph, Molecule\end{tabular} &
  \begin{tabular}[c]{@{}l@{}}Node classification, \\ Knowledge graph reasoning,\\ Graph classification\end{tabular} \\
  &
  \textsc{LLaGA~\cite{chen2024llaga}} &
  \begin{tabular}[c]{@{}l@{}}Text-attributed graph\end{tabular} &
  \begin{tabular}[c]{@{}l@{}}Node classification, \\ Link Prediction, \\ Graph classification\end{tabular} \\
  \midrule
\multirow{6}{*}[-0.8em]{\textbf{\begin{tabular}[c]{@{}l@{}}Domain-specific \\ GFM\end{tabular}}} &
  \textsc{DIG~\cite{zheng2023predicting}} &
  Molecule &
  \begin{tabular}[c]{@{}l@{}}Molecular sampling,\\ Property-guided structure generation.\end{tabular} \\
 &
  \textsc{MACE-MP-0~\cite{batatia2023macemp}} &
  Material Science &
  \begin{tabular}[c]{@{}l@{}}Property predictions of solids, \\ liquids, gases, and chemical reactions.\end{tabular} \\
 &
  \textsc{JMP-1~\cite{shoghi2023jmp}} &
  Material Science &
  Atomic property prediction \\
 &
  \textsc{DPA-2~\cite{zhang2023dpa2}} &
  Material Science &
  Molecular simulation \\
 &
  \textsc{MoleBERT~\cite{xia2023molebert}} &
  Molecule &
  Molecule property prediction \\ \midrule
\multirow{5}{*}[0em]{\textbf{\begin{tabular}[c]{@{}l@{}}Task-specific \\ GFM\end{tabular}}} &
% \textbf{\begin{tabular}[c]{@{}l@{}}Task-specific \\ GFM\end{tabular}} &
  \textsc{ULTRA~\cite{galkin2023towardsULTRA}} &
  Knowledge graph &
  Knowledge graph reasoning \\
  & 
  \textsc{ULTRAQUERY~\cite{galkin2024zero}} &
  Knowledge graph &
  Knowledge graph reasoning \\
  & \textsc{Triplet-GMPNN~\cite{ibarz2022generalist}} 
 & General graph & algorithm reasoning \\ 
 &
  \textsc{G-Retriever~\cite{he2024g}} &
  General graph &
  Graph Question Answer\\  
 &
  \textsc{GraphToken~\cite{perozzi2024let}} &
  General graph &
  Graph Question Answer\\ 
 \bottomrule
\end{tabular}}
\end{table}

\section{Practical Recipes for GFM Applications \label{app:engineer}}
We primarily emphasize the graph principles revolving on transferability and neural scaling law in Section~\ref{sec:principle}, and~\ref{sec:neural-scaling}, respectively. 
In this section, we provide a comprehensive application discussion with more technical details. 
Specifically, we introduce the feature heterogeneity issue, pretext task design, and efficiency issues in subgraph-based methods in Appendix~\ref{subapp:feature-hete},~\ref{subapp:pretext}, and \ref{subapp:subgraph}, respectively.

\subsection{Tackling the Feature Heterogeneity Issue} \label{subapp:feature-hete} 
Existing graph datasets cannot be uniformly utilized for pre-training due to the feature heterogeneity issue induced by missing features or different semantic spaces. 
Feature imputation techniques~\cite{taguchi2021graph, um2023confidencebased, gupta2023grafenne} are generally adapted to predict the missing attributes based on neighboring features. 
However, those techniques require each feature dimension to share the same semantic meaning.
When features are from different semantic spaces, OFA~\cite{oneforall} manually converts the original features with text descriptions and then encodes the embedding with LLMs. 
\citet{oneforall} demonstrates the effectiveness and generality of using LLM embeddings to align heterogeneous node features in the text space. First, it shows that a large portion of feature heterogeneity is caused by the feature engineering process. For example, encoding text using Word2Vec (e.g. OGBN-Arxiv) and TF-IDF (e.g. Pubmed) results in different feature dimensions, leading to heterogeneity. 
If a unified LLM is utilized for encoding, feature heterogeneity can be solved. 
Second, for attributes without text attributes, OFA leverages multi-modal models to project them into textual descriptions. 
Specifically, OFA utilizes GIMLET~\cite{zhao2024gimlet} to generate high-quality text descriptions for chemical molecules, and preliminarily shows that positive transferring can be achieved across diverse domains like text-attributed graphs, knowledge graphs, and molecule graphs after projecting heterogeneous features into text space. 
However, LLM embeddings still have limitations and their performance highly depends on the prompts provided to the LLM text encoder, remaining ample room for exploration in this area. 
One potential way is to borrow ideas from the CV domain. 
\citet{yu2023language} shows that after using LLMs to unify the feature space, further using discrete tokenization for the image can create a better latent space to further improve performance. 

Feature misalignment can also be found in the inference stage between the pre-training model input and the test data. 
\citet{jing2023deepGR} concatenates a learnable padding feature on the downstream task feature to align with the pre-trained GNN. 
However, such a technique cannot adapt to the case when the feature space is not aligned. 
\citet{zhao2023graphtext} directly abandons the original feature and utilizes the feature similarity as guidance.

\subsection{Pretext Task Design}
\label{subapp:pretext}

Given the scarcity of labeled data, a pretext task that can effectively utilize unsupervised data is the cornerstone for larger-scale neural scaling. We provide a brief review of the representative pretext designs. 

Graph contrastive learning designs the pretext tasks~\cite{sun2019infograph, velivckovic2018deepDGI, icml2020_1971MVGRL, you2020graphGCL} to obtain the equivalence via contrasting original and augmented views of the graph without materially changing the semantic content of the input. 
An initial unified understanding~\cite{liu2022revisitingSGCL} on those pre-text tasks illustrates that existing pretext tasks focus on preserving the invariance with the low frequency on the graph spectrum.
Nonetheless, different pretext tasks remain different where \citet{zhu2021empirical} observes that satisfactory performance requires pretext tasks and downstream tasks share similar philosophies, such as homophily. To obtain a pre-training model that benefits different downstream tasks, \citet{ju2023multiParetoGNN} adaptively combined pretext tasks with different philosophies via a multi-task learning framework.

The generative self-supervised learning designs the pretext tasks~\cite{hou2022graphmae, hu2019strategiesAttrMask, kipf2016variationalVGAE} to capture the shared data generation process among different tasks. 
Particularly, they attempt to predict the masked portions of the graph using the remaining structure and features. 
\citet{liu2023rethinkingSIMSGT, xia2023molebert} further observe that task granularity also plays an important role in generative modeling. 
Specifically, employing node-level pretext tasks may lead the model to learn only low-level features~\cite{liu2023rethinkingSIMSGT} while ignoring the global information essential for graph-level tasks.
To address this issue, they adopt a GNN-based tokenizer to explicitly model high-level information in the pre-training stage and thus improve the downstream task performance.

More recently, the next token prediction~(NTP) pretext task~\cite {zhao2023graphgpt} achieves initial success in the molecular graph. 
Notably, this is the first pretext task demonstrating empirical evidence of model scaling.
The potential reason for its success may be (1) the construction of a fixed token set, narrowing down the problem space in a finite set to only predict a discrete token and (2) choosing transformers as the backbone model. 
However, it remains unclear whether the success can be easily extended to more tasks.

\subsection{Efficiency Issues in Subgraph-based Methods.\label{subapp:subgraph}} 
Subgraph-based extraction is a widely adopted technique in GFM to achieve inductive inference~\cite{shaDow} and unify different task formulations~\cite{sun2023all, oneforall}. 
Nonetheless, the subgraph-based extraction leads to the following issues: 
(1) information loss in high-order neighborhoods;
(2) duplicate sub-graph information with excessive memory consumption;
(3) the time complexity of vanilla subgraph extraction grows exponentially with the number of hops, and
(4) the online sub-graph sampling on the fly is also in non-acceptable inference latency~\cite{yin2022algorithm}.

Moreover, the subgraph-based method will increase the number of forward processes for a link-level task, leading to limited efficiency. Typically, for each node pair, we will extract a sub-graph based on them, and apply the forward process. Therefore, the number of forwards increases from $O(|N|)$ to $O(|E|)$, where $|N|$ and $|E|$ are the number of nodes and the number of edges. In practice, the subgraph-based method like SEAL[1] cannot be directly applied to the larger OGB-graph due to such efficiency issues. 
Those issues hinder the applicability of subgraph-based methods.

Designing an effective and efficient sampling method remains a major challenge in building GFM. 
Graph sampling techniques like~\cite{zeng2019graphsaint} and global state vectors~\cite{fey2021gnnautoscale} can help to alleviate these issues. 

\begin{compactenum}
    \item Existing GFM such as PRODIGY~\cite{huang2023prodigy} and OneForAll~\cite{oneforall} based on a subgraph-based view suffers from severe efficiency issues, especially on the link-level tasks. (1) For each node pair, those methods will extract a subgraph based on them, and apply the forward process. Therefore, the number of forwards increases from $O(|N|)$ to $O(|E|)$, where $|N|$ and $|E|$ are the number of nodes and the number of edges. (2) Moreover, the sampling subgraph may also introduce an efficiency problem~\cite{yin2022algorithm}. Subgraph-based methods sample subgraphs in either an offline or online manner. For offline sampling, they need to store subgraph patches for all possible queries, which introduces enormous memory overhead for large graphs. For online sampling, it samples subgraphs on the fly and results in non-acceptable inference latency. 
    \item To solve these efficiency issues, a potential approach is to convert GNN computing to feature precomputation~\cite{BUDDY}. This works for node-level and link-level tasks, but extending it to graph-level tasks is still challenging. 
\end{compactenum}

\section{Additional Principles \label{app:principle}}

\subsection{Principles on deeper GNN design}
In section~\ref{sec:principle}, we emphasize principles revolving on the transferability across datasets. 
Besides, another line of principles focuses on tackling the model limitation towards building effectiveness deeper GNN to capture higher-order structural information.

Principles can tell why vanilla GNNs suffer from performance degradation when increasing the number of layers and provide guidance for solutions. 
The principles can be majorly categorized into the following three perspectives: 
(i) The over-squashing problem~\cite{topping2021understanding} illustrates that the node representation is insensitive to information from important but distant nodes.
(ii) The over-smoothing problem~\cite{oono2019graph, cai2020note} illustrates that more aggregations lead to the node representations converging to a unique equilibrium, which loses the distinction between different nodes. 
(iii) The underreaching~\citep{barcelo2020} illustrates the failure to explore, cover, or affect all relevant nodes in the graph, leading to information loss.
Various techniques are proposed to identify the root causes~\cite{di2023does, wu2023demystifying} and solve the expressiveness issues via new GNN~\cite{yang2021graph} and graph transformer~\cite{nodeformer,muller2023attending} architecture designs. 

Despite those principles are well-studied, they can have a different position and challenges when moving from end-to-end training GNNs to the GFM requiring models to apply across different tasks and datasets. 
Instead of only emphasizing the effectiveness on a single dataset, building GFM raises a novel challenge for us to get good performance with a unified model backbone on diverse datasets. 
The GNN backbone should be able to simultaneously capture discriminative low-order neighborhood information for homophily graphs and high-order neighborhood information for heterophily graphs while the low-order ones may be noisy. 
Current GFMs like OneForAll~\cite{oneforall} empirical solve such a problem via adding virtual nodes with proper prompt designs. 
Nonetheless, there remains a gap between building effective and adaptive deeper GNN for GFM and the current theoretical principles.

\subsection{Additional Description on the Relational Graph Vocabulary of ULTRA}
Typically, the relational vocabulary of ULTRA~\cite{galkin2023towardsULTRA} is inspired by the graph expressiveness theory in~\cite{gao2023double}. 
A concrete example of the relation representation can be found in Figure 2(a) in~\cite{galkin2023towardsULTRA}. 
The relation vocabulary will provide the same embedding for the following two subgraphs with the same relational structure. 
$\textit{Michael Jackson} \xrightarrow[]{\textit{authored}} \textit{Thriller} \xrightarrow[]{\textit{genre}} \textit{disco}$ seamlessly transfers to new entities $\textit{Beatles}  \xrightarrow[]{\textit{authored}} \textit{Let It Be} \xrightarrow[]{\textit{genre}} \textit{rock}$ at inference time. They have the same relational structure with invariant representations regardless of permutations on different node types. 
Interpreting with the graph vocabulary perspective, those two subgraphs should be mapped into the same token. 

Whether the relational vocabulary is suitable or not is according to the graph expressiveness theory. Typically, if two subgraphs are invariant with the same relational structure, i.e., isomorphic to the node type permutation, they will be mapping into the same token with the same representation. In contrast, if two nodes are not invariant, i.e., non-isomorphic to the node type permutation, they will be mapped into different tokens with different representations. 
Overall, the criterion for relational vocabulary is that two sub-structures can be mapped into the same token if and only if two sub-structures are isomorphic.

\section{Discussions \& Open questions}
\label{app:discussion}

\subsection{More discussions on LLMs and Graphs \label{subapp:llm-graph}}

In this section, we provide an extended discussion on leveraging LLMs for graph-related tasks, building on the concepts introduced in Section \ref{subsec:scaling-llm}. 
We provide a more comprehensive discussion on the interaction between LLMs and Graphs.

Specifically, the current integration of graph and foundational models follows two primary pathways. The first involves using graphs to augment the capabilities of other foundational models. The second employs foundational models to address challenges encountered in graph machine learning. 
The first type of work focuses on enhancing the capabilities of foundation models by graphs. \citet{yasunaga2022dragon, yasunaga2022linkbert, jin2023patton, xie2023graphGALM} further pre-train LLMs on text-attributed graphs with a structure-aware pretext task. 
For example, \citet{yasunaga2022linkbert} trains an LLM to predict masked edges, which is formalized as pair classification on two end nodes' attributes. 
Structure-aware training can effectively enhance language models' capability on those tasks requiring structure reasoning, such as multi-hop reasoning. 
These works still view the graph as a second-class citizen providing auxiliary information and put more emphasis on text-centric tasks like question answering~\cite{yasunaga2022dragon, yasunaga2022linkbert}.

The second line of work adopts LLMs' capabilities to solve challenges in the graph domain. \citet{luo2024graphinstruct, chen2024graphwiz, wang2024instructgraph, li2024zerog, ye2024language} utilize the instruction fine-tuning the LLM for various capabilities including zero-shot~\cite{li2024zerog}, link prediction~\cite{ye2024language}, graph reasoning~\cite{luo2024graphinstruct, chen2024graphwiz, wang2024instructgraph}. 
Surprisingly, \citet{wang2024instructgraph} observes that graph fine-tuning can even help those tasks irrelevant with graphs, e.g., mitigate hallucination, logic reasoning, and question answering. 
LLMs can also be utilized for graph generation~\cite{yao2024exploring, wang2024microstructures} where \citet{wang2024microstructures} finds that the graph generated by LLMs is biased towards more triangles and alternating 2-paths, leading to worse performance on the graph recall task.

\textbf{Potential drawback on GNN-enhenced LLM} 
Although these models can perform well, they still have two shortcomings: 
(1) The ability to process structures is bounded by the capabilities of GNN; 
(2) The instruction tuning can be costly while the tuned model can only tackle the corresponding downstream task and is not transferable to other tasks and datasets, which makes their capabilities distant from a GFM.  
We agree that LLM illustrates superior performance on textual node feature understanding.
Nonetheless, it remains unclear whether LLM should play a key role in building GFM or just serve as a better textual feature encoder. 
Moreover, \citet{stechly2023gpt} observes that LLMs are bad at solving graph coloring instances even with multiple-round prompt. 
\citet{yue2023llamarec} points out the efficiency issue of utilizing LLM on the recommendation, the downstream link prediction task. 
The effectiveness and efficiency of LLMs remains unclear.

\subsection{Whether There exists a General Graph Vocabulary?}
A shared graph vocabulary that is effective in transferability across domains and tasks remains an open question. 
In the current stage, we do not speculate the most ideal form of such vocabulary both across tasks and domains. Instead of transferring both across tasks and domains, the current vocabulary design can either transfer across tasks or domains. There is no unified graph vocabulary design at the current research state, as most of the graph vocabulary is either task or domain-specific, e.g., relational vocabulary in ULTRA~\cite{galkin2023towardsULTRA}.

Despite the general graph vocabulary is challenging and not yet realized, we want to introduce one potential way toward it via graph tokenizer training, which is proposed in VQGraph~\cite{yang2023vqgraph}. 
Specifically, it tokenizes nodes with similar structural properties into discrete codes using variants of VQ-VAE~\cite{van2017neural}. After pre-training the tokenizer with a graph reconstruction objective, the discrete codes contained in the codebooks can represent typical structural patterns. The properties of the learned codes will be based on two factors: (1) the encoder and decoder architecture; and (2) the pre-training objective. The current design is still under the guidance of graph principles, remaining not generalized across all the graphs. 
We leave the open question whether there is a universal structure space on graph as the future work

\textbf{Is it possible for GFM to transfer across different domains?} 
For instance, can a model trained on molecular data positively transfer to KG data? 
The answer to this question is initially yes, where we utilize the OneForAll model~\cite{oneforall} as a successful showcase on cross-domain transferring. 
The OneForAll model unifies feature spaces from different domains by using LLM embeddings, map features, and labels into a unified text space with better transferability. 
Such a unified space thereby serves as the basis for GFM that can transfer across citation networks, Wikipedia knowledge graphs, and molecular graphs. 
In the zero-shot setting, OneForAll shows that models trained on citation networks with the node classification task can show positive transfer on molecular graphs, even surpassing the performance on foundation models specific for the science domain like Galactica~\cite{taylor2022galactica}. 
We hypothesize the potential reason is the existence of transferrable patterns among domains, e.g., shared motifs. 
Such transferrable patterns could be modeled by the ability to recognize cycles. For example, 6-cycles are seen in molecules while 3-cycles~(triangles) are critical for social networks~\cite{granovetter1973strength}. 
Moreover, \cite{ribeiro2009strategies} indicates that there are shared patterns between the electronic circuit, the transcriptional network, and the social network despite a severe domain shift. 
More investigations are needed to verify whether models can effectively utilize those transferrable patterns.
For the transferability between knowledge graphs and molecular graphs, we do not have empirical evidence so far. 
We hypothesize that if the knowledge graph involves chemistry-related knowledge, positive transferability can be achievable.

\subsection{Deeper GNNs as the Backbone of GFMs\label{subapp:discuss-deep}}

The development of deep learning generally believes the benefit from deeper Neural Network~\cite{he2016deep}, where the worst case of deeper Neural Networks should be degraded to a shallow solution. 
Notably, we want to emphasize the difference between the general deeper Neural Network design and the deeper GNN design. 
In general deeper Neural Network design, e.g., ResNet~\cite{he2016deep}, Transformer~\cite{vaswani2017attention}, a deeper Neural Network naturally leads to larger parameter scaling. 
However, a deeper GNN does not necessarily lead to larger parameter scaling. 
The key reason is that the GNN is composed of two different components including (1) the feature transformation layer and (2) the aggregation layer. 
Many deeper GNNs focus on increasing the non-parametric aggregation layer while the number of feature transformation layers remains small. 
For instance, the APPNP~\cite{Klicpera2018PredictTP} on Planetoid datasets generally only utilizes two feature transformation layers with a number of parameters less than 10,000. 
It remains skeptical whether deeper GNNs with careful aggregation function design can achieve similar success in other domains without scaling parameter size.

\subsection{Is Invariance a Necessarity for Building GFM?\label{subapp:discuss-invariance}} 
We propose a graph vocabulary perspective emphasizing the invariance among graphs is essential for building Graph Foundation Model. 
However, it remains a mystery whether we should implicitly keep such invariance via specific Message Passing Neural Network~(MPNN) design with equivariance. 
On the one hand, \cite{galkin2023towardsULTRA, galkin2024zero} indicates the effectiveness of building GFM with equivariance. 
On the other hand, \cite{abramson2024accurate, wang2023generating} finds that it is unnesserary to ensure invariance or equivariance with respect to global rotations and translation of the molecule. 
Instead, data argumentation with random rotating and translating is utilized as an implicit regularization during training.
\cite{perozzi2024let}, which first encodes graph with GNNs to conduct prompts and then utilizes LLM for prediction, observes that better performance when breaking the necessary equivariance. 
So far there is no agree on how to preserve geometric equivariance while LLMs also demonstrate potential . 
In additional to geometric Neural Network design, data augmentation, loss functions, and the potential expressiveness of LLM may also provide effective solution.

\subsection{Comparison with Past Relevant Literature. \label{subapp:literature}} 
Concurrent to our position paper, \citet{jin2023largesurvey, li2023surveyGraphLLM, zhang2023graph} reviews those methods adapting large language model~(LLM) for graph, which haven't shown transferring capabilities and thus diverge from our scope to build a graph-centric GFM. 
\citet{liu2023towardsGFMSurvey} further discusses existing graph pre-training and adaption techniques with a focus on their implementations.
Instead of technical details, our work focuses more on the fundamental principles, e.g., geometric invariance across datasets. 
With principle guidance, we depict the promising and relatively elusive directions for the development of GFMs.

\end{document}